%% file: miccai_submission.tex
\documentclass[runningheads]{llncs}

\usepackage[T1]{fontenc}
\usepackage{graphicx}
\usepackage{amsmath}
\usepackage[labelformat=simple]{subcaption}
\usepackage{framed,multirow}

\begin{document}
\title{A training regime to learn unified representations from complementary breast imaging modalities}
%
%
\author{Umang Sharma \and 
Jungkyu Park  \and
Laura Heacock \and
Sumit Chopra  \and 
Krzysztof Geras}


\authorrunning{Anon et al.}
\institute{Anonymous Org
\email{anony@anony.com}} 
\maketitle         
%
%
%
\begin{abstract}
Full Field Digital Mammograms (FFDMs) and Digital Breast Tomosynthesis (DBT) are the two most widely used imaging modalities for breast cancer screening. 
Although DBT has increased cancer detection compared to FFDM, its widespread adoption in clinical practice has been slowed by increased interpretation times and a perceived decrease in the conspicuity of specific lesion types. 
Specifically, the non-inferiority of DBT for microcalcifications remains under debate. 
Due to concerns about the decrease in visual acuity, combined DBT-FFDM acquisitions remain popular, leading to overall increased exam times and radiation dosage. 
Enabling DBT to provide diagnostic information present in both FFDM and DBT would reduce reliance on FFDM, resulting in a reduction in both quantities. 
We propose a machine learning methodology that learns high-level representations leveraging the complementary diagnostic signal from both DBT and FFDM. Experiments on a large-scale data set validate our claims and show that our representations  enable more accurate breast lesion detection than any DBT- or FFDM-based model. 

\end{abstract}
\input{Sections/introduction}
\input{Sections/background}
\input{Sections/motivation}
\input{Sections/methods}
\input{Sections/experiments}

\input{Sections/discussion}
\begin{credits}
\subsubsection{\ackname} 
This work was supported in part by grants from anonymous institution (******), anonymous institution (******) and anonymous institution (******)

\subsubsection{\discintname}
The authors have no competing interests to declare that are
relevant to the content of this article. 
\end{credits}
%
%
%
\newpage
\bibliographystyle{splncs04}
\bibliography{miccai_submission}
%




\input{Sections/supplementary}
\end{document}

%% file: Sections/introduction.tex
\section{Introduction}
\label{sec1}
\begin{figure}[t!]
  \centering
{{\includegraphics[width=0.45\textwidth]{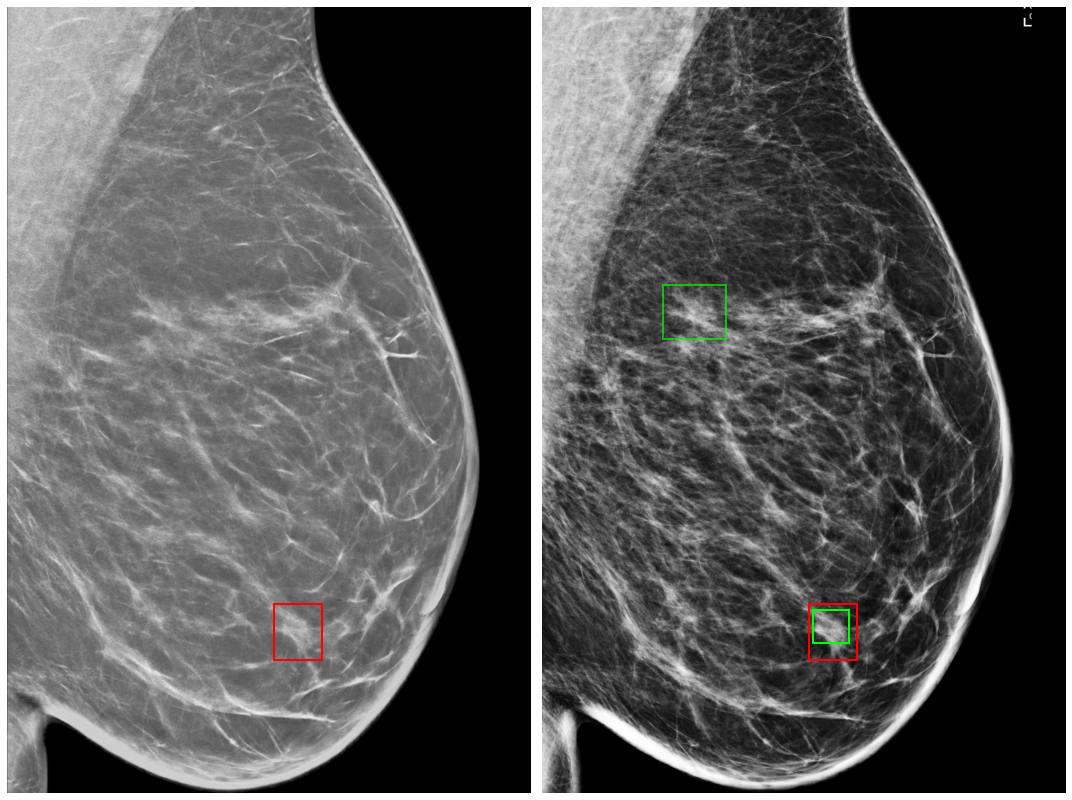} }}%
   \qquad 
{{\includegraphics[width=0.45\textwidth]{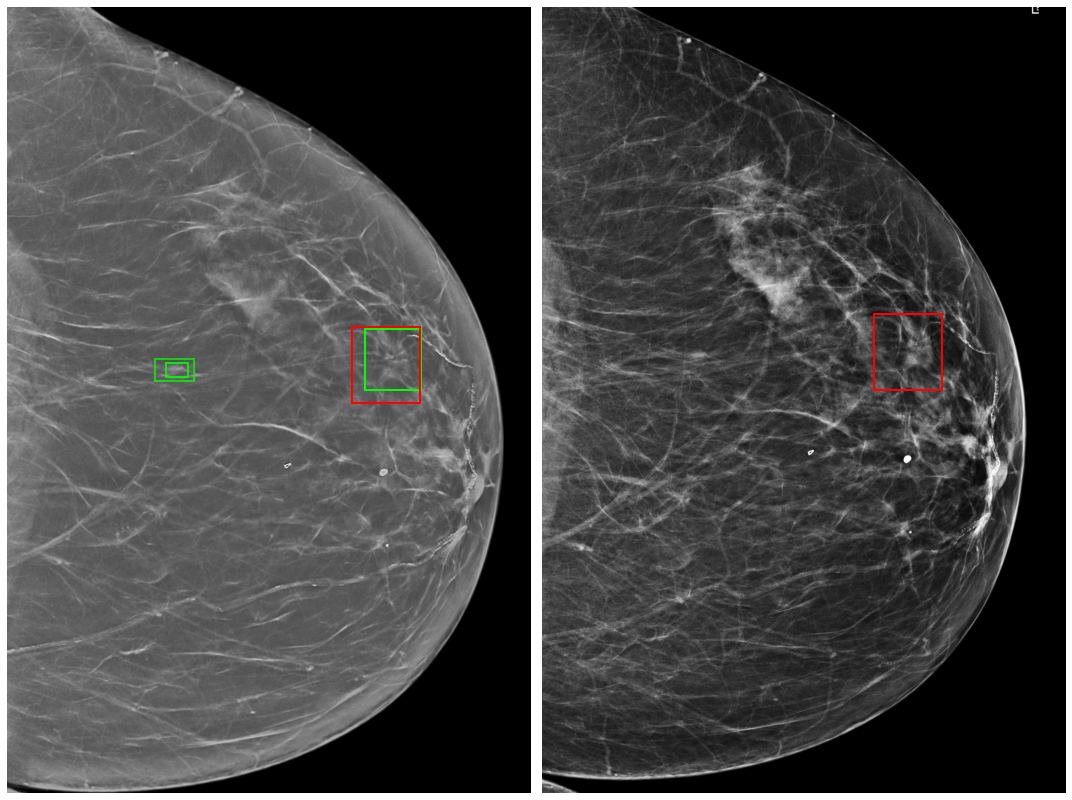} }}%
\caption{Examples of complementary information in FFDMs and SMs. 
In each pair, the right image is FFDM and the left is the corresponding SM.  
Red and green boxes are lesions marked by an expert and a model, respectively. 
A model using FFDM and a model using SM capture different sets of lesions. 
}
\label{fig:comp_info}
\end{figure}

The standard of care in breast cancer screening is 2D imaging of the breast using Full Field Digital Mammography (FFDM). 
More recently, Digital Breast Tomosynthesis (DBT), which captures multiple slices to create a 3D view of the breast, was developed to improve accuracy
without significant change to the radiation dosage \cite{dbt_review}. 
Although DBT slices capture a more accurate view of breast tissue, 
interpreting the 3D slices individually can limit the evaluation of regional or global patterns, including microcalcifications. 
To circumvent this concern, a 2D synthetic mammogram (SM), designed to capture whole breast information similar to a typical FFDM image, is created from the DBT slices. 
It is standard practice to accompany DBT slices with an SM. 
While SMs expedite image interpretation, they are known to produce artifacts \cite{proconsms,proconsms2}. 
For instance, small calcifications ($<$ 2 cm) are more challenging to classify as malignant or benign in SMs because of contrast enhancement to make suspicious findings stand out. This also increases the conspicuity of benign calcifications compared to FFDMs. 
On the other hand, SMs 
can emphasize lesions hidden by overlying tissue structures, while these lesions may be obscured by dense tissue on FFDMs (see Figure \ref{fig:comp_info}). 
Such complementary visual information has led to use of combined acquisition of DBT and FFDM instead of DBT alone, increasing screening times and radiation dosage and thus impacting patient care.
With the goal to decrease screening time and radiation dosage, we propose a novel machine learning model that could reduce reliance on FFDM in the presence of DBT+SM. 
We hypothesize that if we have access to a large collection of pairs of FFDM and the corresponding DBT+SM images during training, we can learn to encode the relationship between the two modalities and leverage this information during inference to make accurate predictions using only a single modality. 
We conjecture that if the relationships are learnt well, then for a given imaging modality the predictions that leverage these learnt relationships will be more accurate than the predictions that do not leverage them. 
Towards that end, we propose a ML model and a training methodology, that learns to capture relationships between SM and FFDM, and uses it to generate high-level representations of SM that also encode knowledge from the corresponding FFDM images, without actually using FFDMs at inference.  
We validate our hypothesis by developing our model for detecting lesions in mammograms. 
Inspired by the recent success of Deep Learning (DL) models in both natural images \cite{maskrcnn,yolo,ssd,resnet} and medical imaging \cite{ribli,geras_moy,cancers15102704,shen2019,yala2019,li2021,wu2020,dbtex_comp_2,dbtex_comp}, we modify the EfficientDet \cite{efficientdet} to learn these representations. 
The results show that our method outperforms models that simply learn and predict using either FFDM or SM alone.
To the best of our knowledge, our work is first-of-its-kind that attempts to capture relationships among multiple modalities and uses the information to improve predictions from a single modality during inference.

%% file: Sections/background.tex
\section{Background and Related Work}
\label{sec-2}

\textbf{Imaging for Breast Cancer Screening: }
\label{sec-2.1}
The 2D Full-Field Digital Mammography (FFDM) and 3D Digital Breast Tomosynthesis (DBT) are the current standard of care in screening for breast cancer at population-level. 
While the 3D volume allows for better visualization and localization of lesions hidden within dense overlapping tissues, the 2D image allows for a more efficient read and better captures structures that are spread across the depth and not obvious when viewed in any single slice. 
The 3D DBT slices are therefore accompanied by a 2D Synthetic Mammogram (SM) image, which is reconstructed from the 3D projections of DBT. 
SM and DBT slices are interpreted in conjunction, allowing both a global and local evaluation of the breast parenchyma. 
This workflow has led to increased cancer detection rates and lower recall rates \cite{ciatto,haas2013,skaane2019}.  
While the hope was for SM to eventually replace FFDM, studies have shown that SMs introduce artifacts, such as pseudo-distortions and pseudo-calcifications \cite{proconsms,proconsms2}, leading to increased recall rates for some subgroups of imaging findings. 
There is also additional debate about the visibility of microcalcifications on SM compared to FFDM \cite{spangler2011,horvat2019,GUR2012166}.
This has led to continued use of combined DBT and FFDM acquisition at many breast imaging centers, which allows for cross-verification against both image types. 
However, this combination protocol increases radiation dose and exam time, causing patient discomfort and potential harm. 

\noindent
\textbf{Deep Learning Models for Image Analysis:}
Multiple deep learning architectures have been proposed for the problem of object detection and localization in natural images. 
These include R-CNN \cite{fastrcnn,fasterrcnn,maskrcnn}, YOLO \cite{yolo,yolov3,yolo9000}, SSD \cite{ssd}, Overfeat \cite{overfeat} and Efficient-Det \cite{efficientdet}. 
For our purpose, we view any DL model as being composed of two modules, namely the ``feature extractor'' and the ``predictor'', stacked on each other. 
The feature extractor, takes the raw data (e.g., images) as input and extracts high-level features (a.k.a., representations), that are invariant to arbitrary artifacts irrelevant to the final task. 
The predictor, takes these high-level representations as input and provides the final answer such as predicted class or bounding boxes. 
In this work we extend the Efficient-Det architecture to extract representations from SM images that contain information from both the SM and the corresponding FFDM images. 

\noindent 
\textbf{Knowledge Distillation in Deep Learning Models:}
\label{sec:2.3}
Knowledge distillation refers to the concept of transferring (distilling) information (or knowledge) from one model to another. 
In \cite{hinton2015distilling}, the authors learn smaller models using the high-level features learned by a larger model. 
Several works have used knowledge distillation for tasks such as diagnosis from different types of MR images \cite{lu2019}, video action recognition \cite{wu2019}, predicting high-resolution image features from low resolution images \cite{zhu2019}. Recently, \cite{sobal2022joint,assran2023selfsupervised} make use of a Joint-Embedding Predictive Architecture (JEPA) to learn common representations between similar inputs that are tailored for a specific task. In this work, we use similar ideas to transfer knowledge learned from FFDM into the representations generated by SM. 

%% file: Sections/motivation.tex
\section{Complimentary Information Between FFDM and SM}
\label{sec:motiv}
\begin{figure*}[t!]
\centering
\includegraphics[width=1\textwidth]{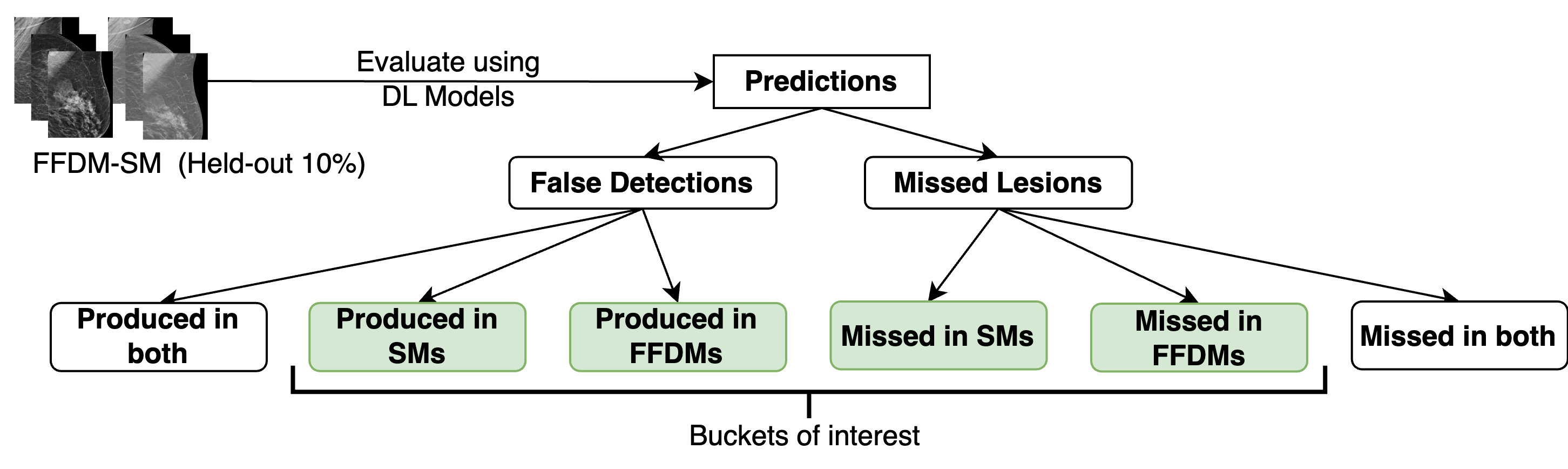}
\caption{Framework to compare FFDMs against SMs for disease detection. 
}
\label{fig:analysis}
\end{figure*}

To validate our hypothesis, that FFDM and SM images contain complementary information, 
we compare the disease detection performance of SMs against FFDMs, on a   
large-scale anonymous Breast Cancer Screening Dataset \cite{nyudbt} which consists of $1,239,372$ pairs of FFDM and SM images (see Table~\ref{tab:dataset-stats}). 
As a surrogate to radiological reads, we use a pair of identical deep learning models, each trained individually on either FFDMs or SMs to localize and classify lesions as malignant or benign. 
For every image, the model outputs  a set of detections accompanied with a confidence score. 
Detections with a confidence score lower than a threshold (chosen by pegging the average false positive detection rate to $1$ per image over the validation set) are filtered out. 
Finally, we consider predictions as true positives if the distance from its center point to the center point of the ground truth is less than half the diagonal of the ground truth bounding-box or 100 pixels whichever is larger~\cite{dbtex_comp,dbtex_comp_2}.
The remaining predictions are grouped into ``Missed Lesions'' (when the predictions miss the ground truth lesion) and ``False Detections'' (when there is no underlying ground truth lesion). 
These are further segregated into 4 buckets of interest as shown in Figure~\ref{fig:analysis}. 
\newline \textbf{Missed in FFDM:}
We expect model using FFDMs to have difficulty visualizing lesions otherwise obscured by dense overlying fibroglandular tissue. 
As such, the images in this bucket involve dense breasts with significant tissue overlap (see supplementary Figure~\ref{fig-ffdm-fn}).
Since the SM images are constructed from the DBT slices, the model using SMs is expected to do a better job of localizing lesions that may be seen only on several internal breast slices. 
\newline \textbf{Missed in SM:} Mammograms in this bucket contain masses lacking distortion or with subtle calcifications. Masses without spiculation or distortion are often emphasized less on SM images. Punctate microcalcifications may similarly be less emphasized and blend in with the surrounding dense tissues (Figure~\ref{fig-cview-fn}).
\newline \textbf{False Detections:}
Since the chosen thresholds for each model allows for $1$ false positive detection per image on average, we end up with a large number of images for which the models falsely predict a lesion, making manual analysis of images in these buckets infeasible. 
We pick a subset of 20 images uniformly at random from each of the ``False Detection'' buckets of interest. 
$8$ out of $20$ images, for which the model trained on SMs produced false detections, comprised of pseudo-distortions (Figure~\ref{fig-cview-fp}(a)) and pseudo-calcifications (Figure~\ref{fig-cview-fp}(b)). 
These observations are consistent with the belief that algorithms used to generate synthetic SM images from the cross-sectional DBT slices make distortion or calcification-like findings more ominous than they would appear on FFDMs \cite{proconsms2}. 
Majority of the false positives produced in FFDMs correspond to high density breasts with significant tissue overlap. Specifically, $14$ out of $20$ images in this bucket had a breast density BI-RADS of 3 (heterogeneously dense) or 4 (extremely dense) (Figure~\ref{fig-ffdm-fp}). In summary, our analysis shows that FFDMs and SMs contain complementary information regarding estimating the presence, or absence, of suspicious lesions. 
In the following sections, this complementary information is used to improve the performance of the models for lesion detection.


%% file: Sections/methods.tex
\section{Learning From Complementary Imaging Modalities}
\begin{figure*}[t!]
\centering
\includegraphics[width=\linewidth]{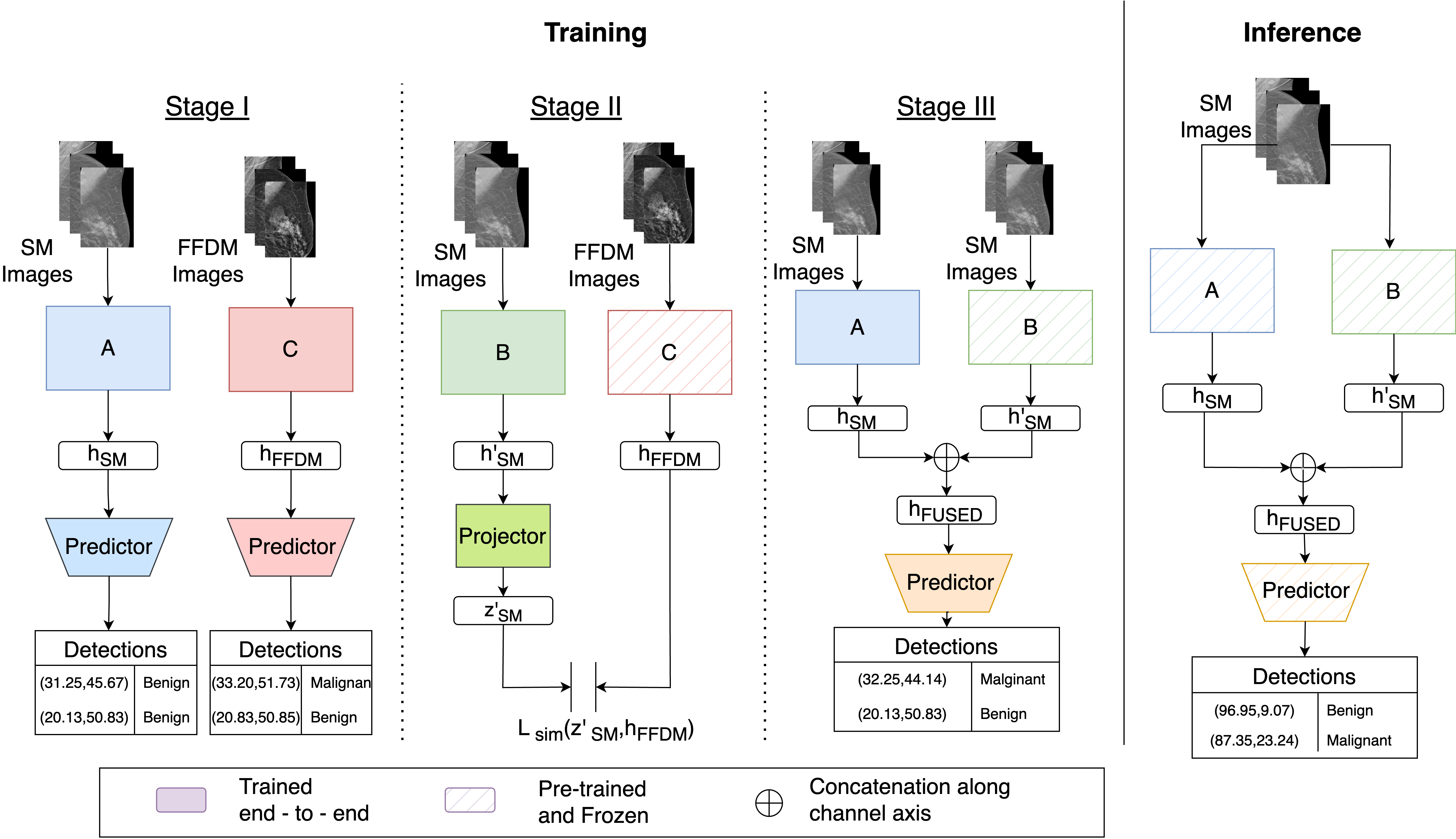}
\caption{Proposed framework to learn representations from SM encoding knowledge from FFDM. 
Training stages: {\bf (I)} A and C are trained individually on SMs and FFDMs; 
{\bf (II)} B is trained to produce representations of SMs that borrow knowledge from FFDMs; 
{\bf (III)}  Fused representations are obtained by concatenating the features from A and B. 
At inference, only SMs are used to generate the fused representations $h_{\textsc{fused}}$ which are then processed to make predictions.}
\label{fig:arch}
\end{figure*}

Based on our analysis in Section \ref{sec:motiv}, 
learning a representation of data that captures information from both the modalities should provide higher predictive power. 
We now describe our proposed methodology that learns representations from SMs that encode complementary knowledge from FFDMs. 
Critically, during inference our methodology only uses SM images to predict the disease. 

\noindent 
\textbf{Model Architecture:}
Our framework consists of 3 modules, denoted by (A), (B), (C) in Figure \ref{fig:arch}. 
Each module is an instantiation of Efficient-Det \cite{efficientdet} whose parameters are pre-trained on Microsoft COCO dataset \cite{coco}. 
This training and inference regimes for these modules is discussed next. 

\noindent 
\textbf{Model Training}
\label{sec:training}
The entire system  is trained in three stages (see Figure ~\ref{fig:arch}). 

\noindent 
\textit{Stage I: } Modules A and C respectively take SM and FFDM images as inputs and are trained to localize the lesions by optimizing the focal loss \cite{focal_loss} ($L_{Det}$): 
\begin{align}
\label{eqn:loss_det}
L_{\text{det}}(p) &= -\alpha (1 - p)^\gamma \log(p).
\end{align}
Here $\alpha$ and $\gamma$ are tunable hyperparameters, while $p$ is the confidence score for a detection. 
The focusing parameter $\gamma$ makes the loss robust to dataset imbalance by enabling to  focus on harder to predict examples. 
Following this, we freeze the learned parameters and remove the final predictor layers such that A and C produce high-level representations denoted by $h_{\textsc{sm}}$ and $h_{\textsc{ffdm}}$, respectively. 

\noindent 
\textit{Stage II:} Module B is trained to mimic the outputs of module C. 
It takes SM images as input and produces a representation $h_{\textsc{sm}}'$. 
The goal of training is to bring $h_{\textsc{sm}}'$ close to $h_{\textsc{ffdm}}$ (the representation derived from FFDMs by C in Stage I). 
This is achieved by first linearly projecting $h_{\textsc{sm}}'$ to generate another feature vector $z_{\textsc{sm}}'$ and then minimizing the negative of cosine similarity loss $L_{sim}(z_{\textsc{sm}}',h_{\textsc{ffdm}})$, where $L_{sim}$ is defined as: 
\begin{align}
\label{eqn:loss_sim}
L_{\text{sim}}(h_i,h_j) &= \frac{h_i \cdot h_j}{\max(\mid\mid h_i \mid\mid_2 \cdot \mid\mid h_j \mid\mid_2, \epsilon)}.
\end{align} 
The linear projection layer applies  a single convolution filter of size $3\times 3$ along with padding to preserve the dimensionality.
Note that in stage II, module C is kept frozen from Stage I, to provide stationary ground truths for training B.
While it might not be possible to recover all the information present in the FFDM image, we argue that learning to mimic a high-level representation can learn some relationships in the pairs of images it sees during training.
In future work we will better quantify the relationships this technique can learn to capture. 

\noindent 
\textit{Stage III:} For an input SM image, we concatenate the outputs from learnt modules A and B to form a fused representation $h_\textsc{fused} = h_\textsc{sm} \oplus h_\textsc{sm}'$ and feed them into the final detection model. 
This detection model comprises of the final predictor layers of the Efficient-Det architecture, which is trained by optimizing $L_{det}$ loss (Equation~\ref{eqn:loss_det}). Furthermore, we use the gradients from this stage to fine-tune the weights of module A, while keeping $h_{\textsc{sm}}'$ fixed.

\noindent 
\textbf{Inference}
Inference involves taking a SM image as input, passing it through modules A and B to generate the fused representations $h_{\textsc{fused}}$ (from $h_{\textsc{sm}}$ and $h_{\textsc{sm}}'$), and subsequently passing it in to the predictor layers. 
We emphasize that the module C, which is trained on FFDM images, is not used at this stage. 

\noindent 
\textbf{Evaluation Metrics}
The performance of our framework is evaluated  using the free-response receiver operating characteristic (FROC) curves \cite{Bandos2009,dbtex_comp,cad_comp}.
It indicates the sensitivity of a model with respect to the average number of false positive predictions made per image. 
The FROC curve defines a metric FAUC-$x$, indicating the area under the FROC curve, where the curve is cut off at $x$ false positives per image on average. 
We report FAUC-$1$ in our experiments. 

%% file: Sections/experiments.tex
\section{Experiments and Results}
\begin{table*}[t!]
\caption{Anonymous Breast Cancer Detection Statistics}
\label{tab:dataset-stats}
\begin{tabular}{l|l|l|l|l|l|l}
 \hline
 Image Statistics & Malig. & Benign & Both & Unannotated & Negative Images & Total \\
 \hline
Lesion Segmentation & 1594 & 2469 & 54 & 24871 & 1,210,384 & 1,239,372 \\
Pathology Report & 2248 & 14410 & 1488 & N/A & 1,221,226 & 1,239,372\\
 \hline
\end{tabular}
\end{table*}
\textbf{Dataset and Hyperparameters: }
We trained and evaluated our model on the anonymous Breast Cancer Screening Dataset \cite{nyudbt} consisting of 1,239,372 FFDM-SM image pairs. 
The FFDM images, which consists of all four views (R-CC, L-CC, R-MLO, and L-MLO) are of size $4096 \times 3328$ or $3328 \times 2560$ pixels, while the SM images are of size  $2457 \times 1996$ or $2457 \times 1890$ pixels. 
We pre-process the images in the same way as described in Section 2B of \cite{nyudbt}.
Our reference standard is derived from the  pathology reports and radiologist readings. 
Each image is associated with a binary class label indicating the presence/absence of a malignant lesion, extracted from pathology reports available as part of \cite{nyudbt}. 
Images with positive findings are accompanied with a pixel-level annotation indicating the position of the lesions annotated by 
a board-certified radiologist.  
$24,871$ images were not examined by a radiologist and hence do not have pixel-level annotations and are excluded from the dataset. 
The remaining studies are divided into three disjoint sets at the patient-level to create training (60\%), validation (10\%), and test (30\%) sets.  
Data imbalance is addressed by using 
a weighted sampling strategy that samples an equal number of positive and negative images per epoch. 
For tuning the hyperparameter ($\alpha$, $\gamma_0, \gamma_1$), we performed a grid search, using the validation set to select the optimal values. 
We find that a learning rate of $1.5e-4$, $\alpha=0.9$, $\gamma = 2.5$ works best for our models.




\begin{table*}[t!]
\begin{center}
\caption{Detection metrics (with 95\% CI) for various models. 
\textsc{FAUC-1 Val+} and \textsc{FAUC-1 Test+} are computed over images with at least one positive finding.}
\label{tab:baseline-results}
\begin{tabular}{l|l|l} 
 \hline
 \textsc{Experiment}&\textsc{FAUC-1 Val+/FAUC-1 Val}&\textsc{FAUC-1 Test+/FAUC-1 Test}\\
 \hline
 \multirow{2}{*} \text{\scshape Model$_{\textsc{sm}}$}& 0.617(0.58-0.65)/0.59(0.55-0.63)&0.523(0.50-0.54)/0.515(0.497-0.533)\\ 
  \textsc{Model$_{\textsc{ffdm}}$}&0.587(0.544-0.63)/0.57(0.54-0.61)&0.54(0.52-0.557)/0.493(0.476-0.51)\\ 
  \hline
 $\textsc{Base}_{\textsc{UB}}$&0.662(0.63-0.703)/0.624(0.61-0.65)&0.59(0.576-0.606)/0.565(0.562-0.585)\\ 
 \hline
 Fused Model&0.656(0.617-0.69)/0.615(0.58-0.66)&0.565(0.550-0.584)/0.538(0.52-0.56)\\ 
 \hline
\end{tabular}
\end{center}
\end{table*}

\noindent 
\textbf{Baselines, Performance Upper Bounds and Results}
We compare the performance of our model against  \textsc{Model$_\textsc{ffdm}$} and \textsc{Model$_\textsc{sm}$}, which are Efficient-Dets trained independently on FFDM and SM images respectively. 
We also establish an upper bound on the best performance any model could achieve if it had access to the true representations extracted from FFDM and SM images. 
This was accomplished by first creating a fused representations ($h_\textsc{fused}$) by concatenating the representations of the FFDM images ($h_\textsc{ffdm}$) with the representations of the SM images ($h_\textsc{sm}$). 
The fused representations were then passed to the downstream detection model, which was fine-tuned end-to-end. 
We denote this upper bound model by \textsc{Base$_\textsc{UB}$}. 
Our results are summarized in Table \ref{tab:baseline-results}. 
We observe that across both splits, our method outperforms either baseline model and approaches the performance of the upper bound model.

%% file: Sections/discussion.tex
 \section{Discussion and Conclusion}
As further analysis, we answer the following questions to assess performance gains on a per-case basis.
\newline \textbf{Does the fused model lead to fewer false negatives compared to a model trained using either modalities?} 
    \textsc{Model$_{\textsc{sm}}$} misses lesions in $78$ images of which $52$ were also missed by \textsc{Model$_{\textsc{ffdm}}$}. 
    Overall \textsc{Model$_{\textsc{ffdm}}$} misses lesions in $86$ images. 
    In comparison, the fused model was able to capture lesions in $18$ images that were missed by 
    \textsc{Model$_{\textsc{sm}}$} and in $10$ images that were missed by both the baseline models. 
    In total, the fused model misses lesions in $66$ images.
    
    \noindent 
    \textbf{Can the fused model detect lesions missed by \textsc{Model$_{\textsc{sm}}$} but captured by \textsc{Model$_{\textsc{ffdm}}$}?} 
    From our analysis in Section~\ref{sec:motiv}, we observe that \textsc{Model$_{\textsc{sm}}$} missed malignant lesions lacking distortions or calcifications. 
    Figure~\ref{fig:image-analysis}{(A)} shows a couple of cases where \textsc{Model$_{\textsc{sm}}$} misses a lesion that \textsc{Model$_{\textsc{ffdm}}$} correctly captures, and the rightmost image shows the fused model able to capture the lesion on the same mammogram without ever seeing the associated FFDM image. 
    In total, the fused model is able to capture lesions on 10 images that were also captured by \textsc{Model$_{\textsc{ffdm}}$} but not by \textsc{Model$_{\textsc{sm}}$}.

    \noindent 
    \textbf{Can the fused model use knowledge distilled during training and perform well on cases where both single modality models fail?}
    Since the fused model estimates representations derived from the FFDM images using the SM images, it is reasonable to assume that it learns some relationships between the representations of the two modalities. 
    Figure~\ref{fig:image-analysis}{(B)} shows that both  \textsc{Model$_{\textsc{ffdm}}$} and \textsc{Model$_{\textsc{sm}}$} miss the ground truth lesions. 
    The third image in each set shows the correct predictions made by the fused model. 
    In total, there were 8 images where neither baseline model was able to capture the lesion but the fused model does.

    \noindent 
    \textbf{Are there cases where the fused model is worse than  \textsc{Model$_{\textsc{sm}}$}?} 
    Since the fused model has additional training signal by mimicking representations of FFDM images, there are cases where it produces incorrect predictions even when the baseline single modality \textsc{Model$_{\textsc{sm}}$} makes a correct prediction. Figure~\ref{fig:image-analysis}{(C)}, shows such samples. Overall, there are 66 cases where the fused model misses a lesion. Out of these, all three models missed in $44$ cases. Among the remaining 22 cases, 3 were correctly classified by \textsc{Model$_{\textsc{sm}}$}, 16 by \textsc{Model$_{\textsc{ffdm}}$} and 3 cases where both baseline models were able to give accurate predictions.
    \newline In conclusion, we present an approach for integrating knowledge across different medical imaging modalities, specifically harnessing the synergistic potential of FFDM and SM. Our methodology effectively enhances lesion detection in mammography by leveraging complementary information from both modalities, while only requiring SM images as input during inference. 
    We believe this setup can be generalized to broader applications in medical imaging.

%% file: Sections/supplementary.tex
\makeatletter
\renewcommand \thesection{S\@arabic\c@section}
\renewcommand \thefigure{S\@arabic\c@figure}

\makeatother
\newpage
\section*{Supplementary Material}

   \begin{figure*}[h!]
    \centering
    \begin{minipage}[b]{\linewidth}
           \centering
        {{\includegraphics[width=0.49\textwidth]{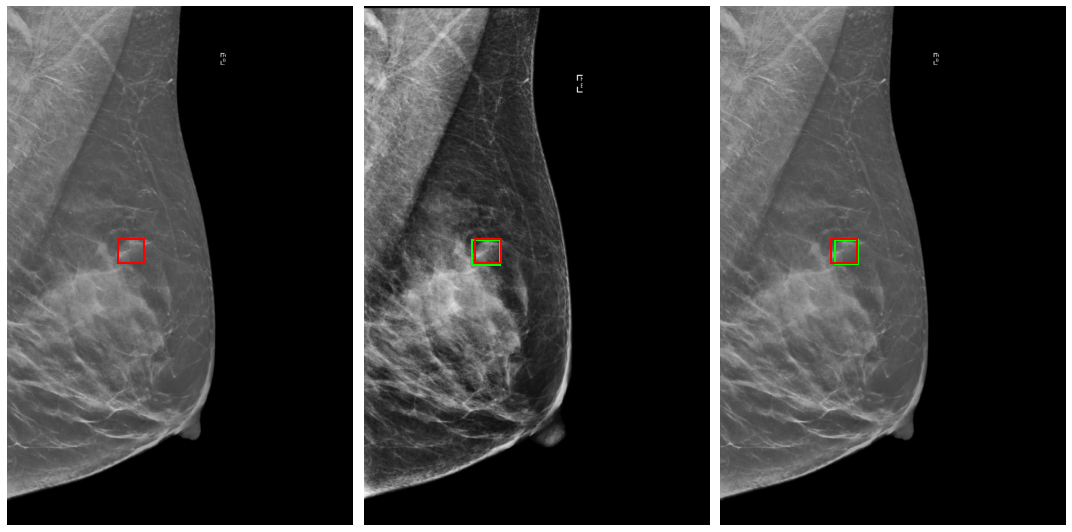} }}
        {{\includegraphics[width=0.49\textwidth]{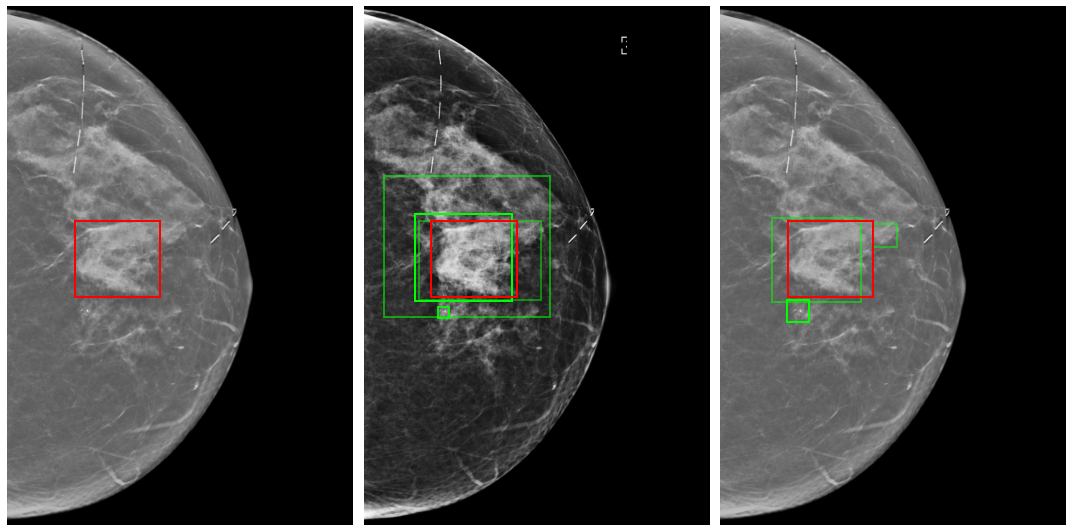} }}
        \caption*{(A) The fused model is able to capture lesions missed by \textsc{Model$_{\textsc{sm}}$}. While \textsc{Model$_{\textsc{ffdm}}$} is able to capture it, the fused model never actually sees the FFDM image.}
    \end{minipage}
    \begin{minipage}[h]{\linewidth}
        \centering
    {{\includegraphics[width=0.49\textwidth]{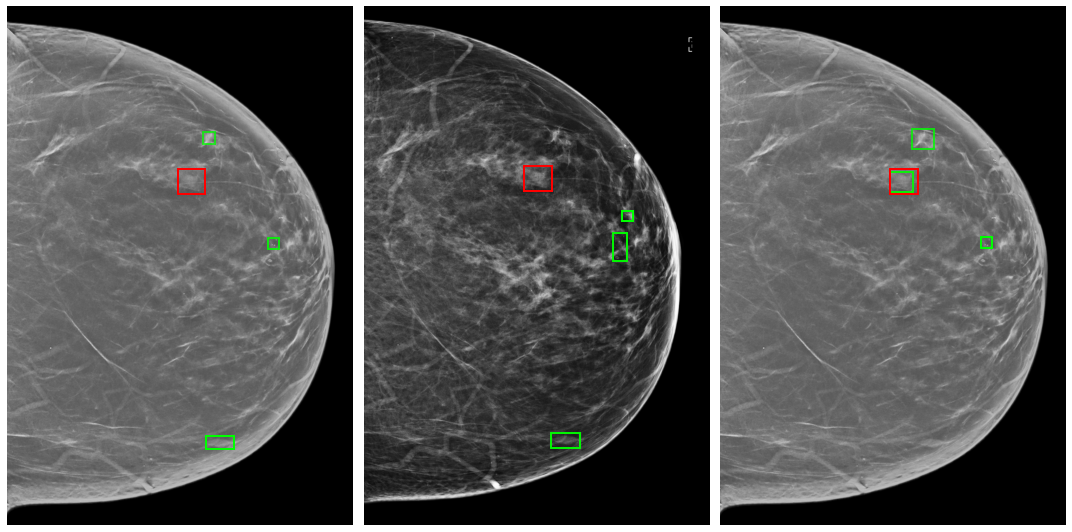} }}
    {{\includegraphics[width=0.49\textwidth]{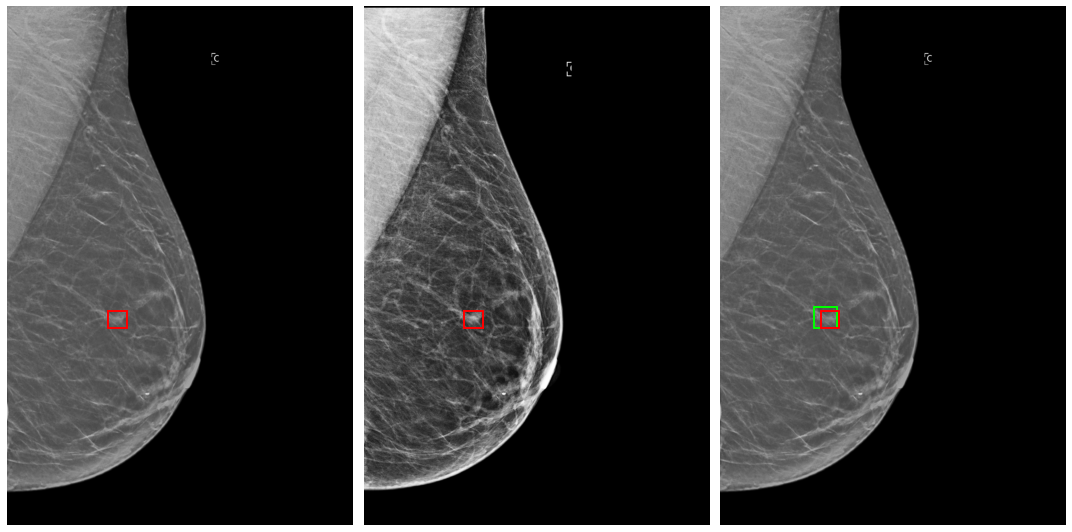} }}
    \caption*{(B) The fused model is able to capture lesions missed by both \textsc{Model$_{\textsc{sm}}$} and \textsc{Model$_{\textsc{ffdm}}$}.}
    \end{minipage} 
    \begin{minipage}[b]{\linewidth}
        \centering
        {{\includegraphics[width=0.49\textwidth]{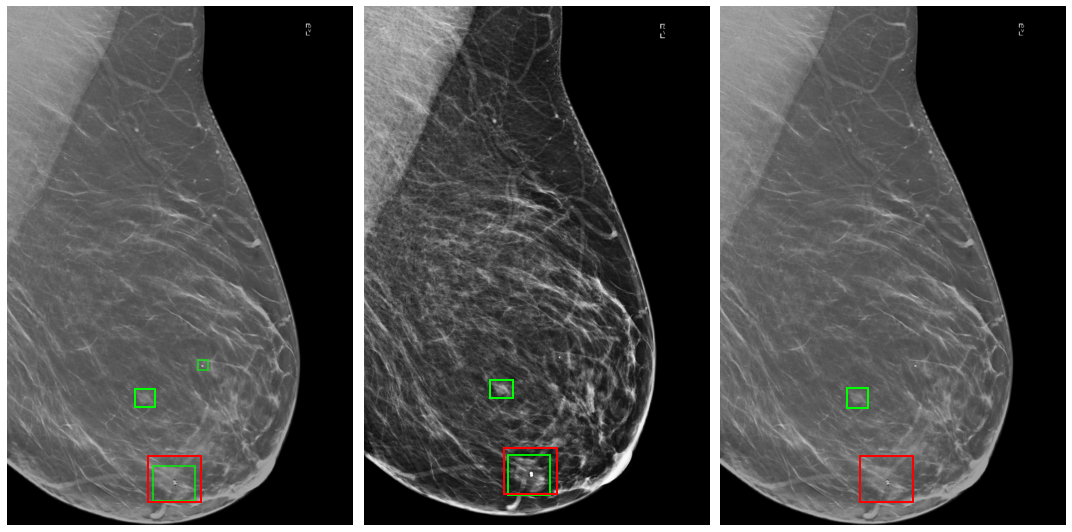} }}
        {{\includegraphics[width=0.49\textwidth]{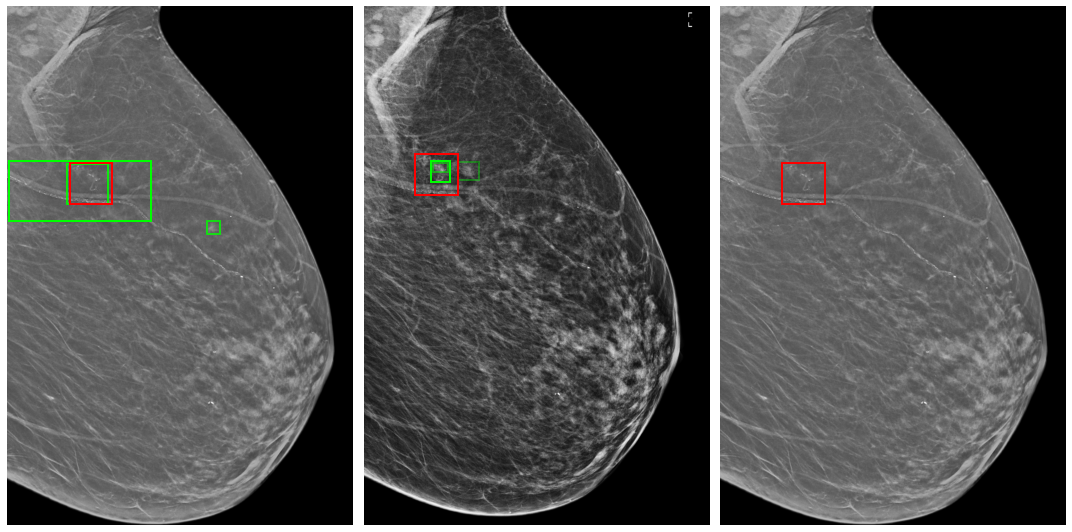} }}
        \caption*{(C) Cases where the fused model (rightmost image in each set) fails to capture the underlying lesion while both \textsc{Model$_{\textsc{sm}}$} and \textsc{Model$_{\textsc{ffdm}}$} are able to localize it.}
    \end{minipage}
    \caption{Each row contains two sets of three images corresponding to predictions made by the SM model (left), FFDM model (center) and the Fused Model (right). Red bounding boxes denote a ground truth lesion marked by an expert, while the green boxes are the predictions made by a trained model.}
    \label{fig:image-analysis}
\end{figure*}

\begin{figure}[!h]
  \centering
  \begin{minipage}[b]{\linewidth}
      {{\includegraphics[width=0.45\textwidth]{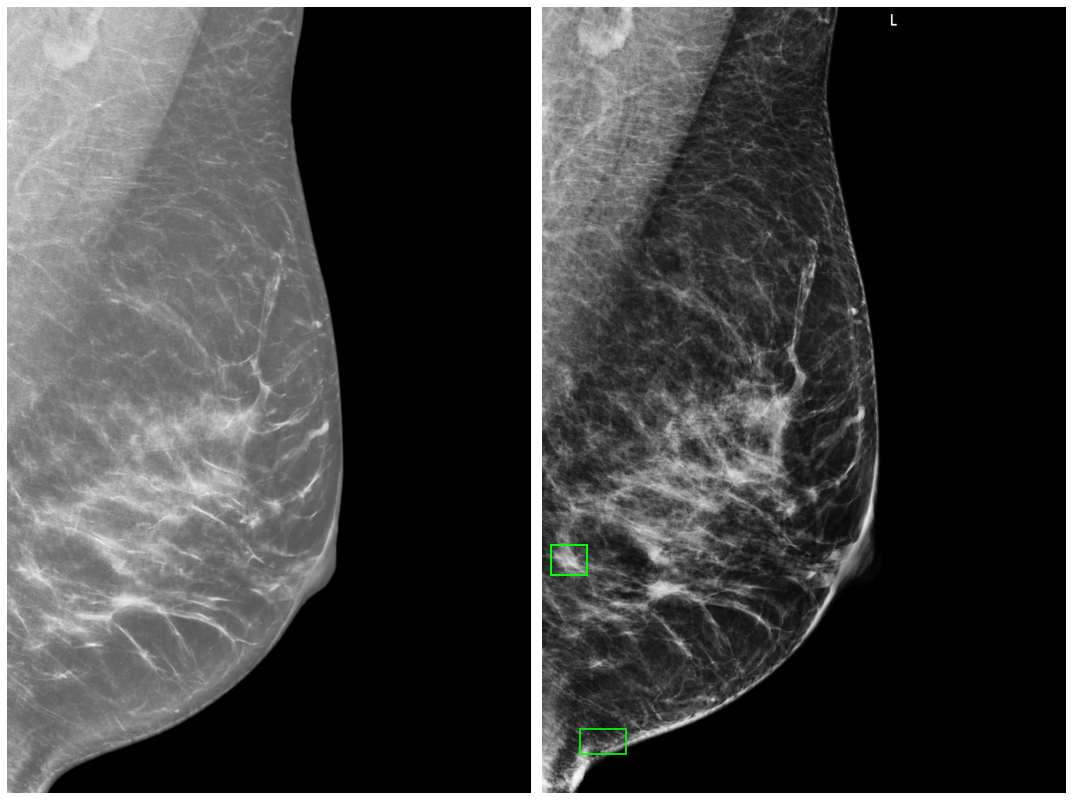} }}%
\qquad
 {{\includegraphics[width=0.45\textwidth]{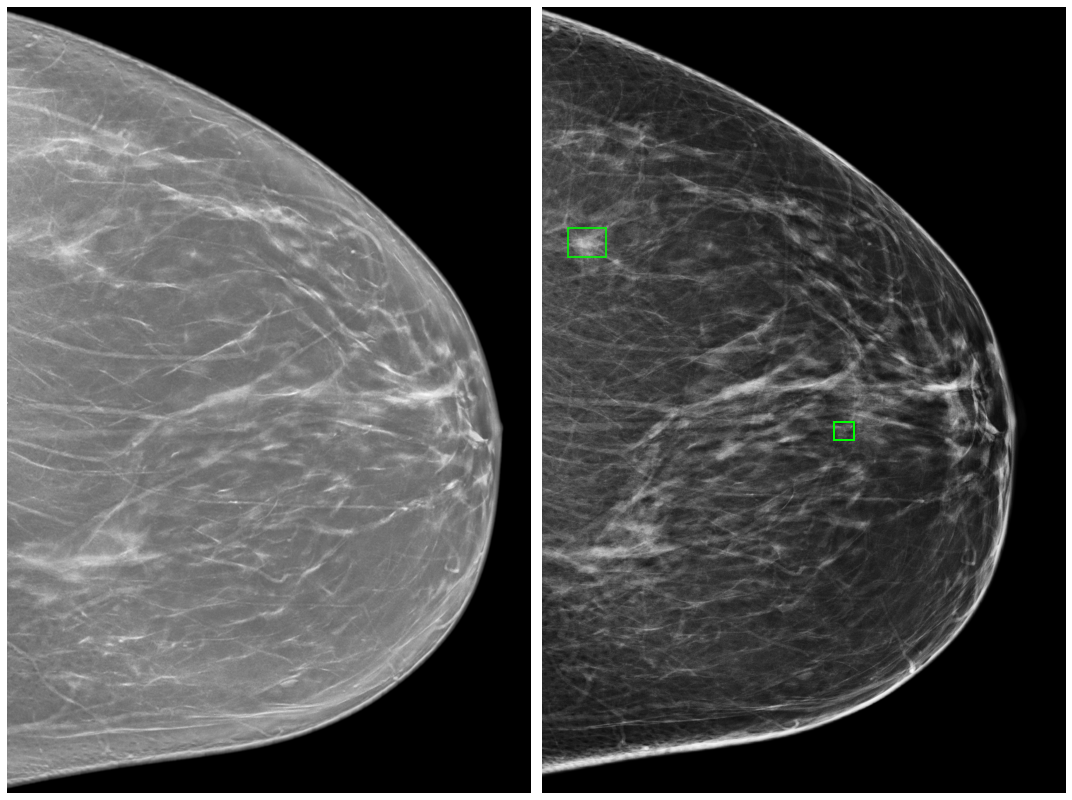} }}%
    \caption{False positive mis-classifications made by the FFDM model only. SM image is on the left of each set, FFDM to the right.}
    \label{fig-ffdm-fp}
  \end{minipage}
  
  \begin{minipage}[h]{\linewidth}
      {{\includegraphics[width=0.45\textwidth]{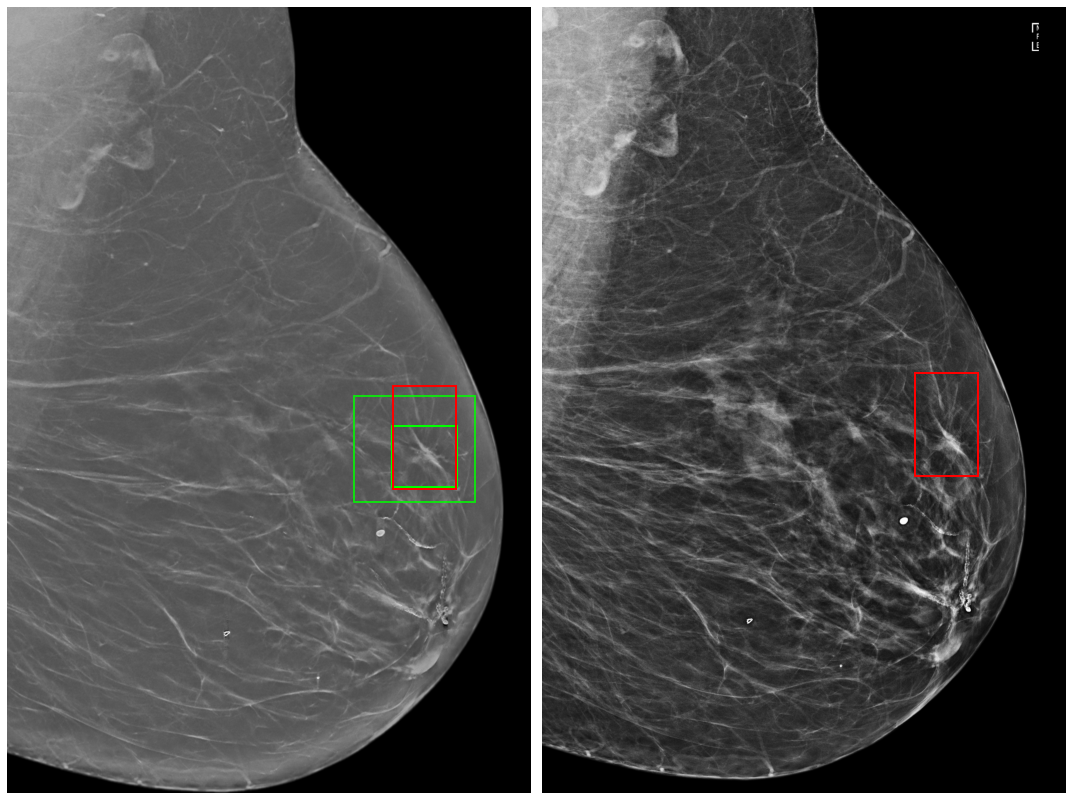} }}%
    \qquad
 {{\includegraphics[width=0.45\textwidth]{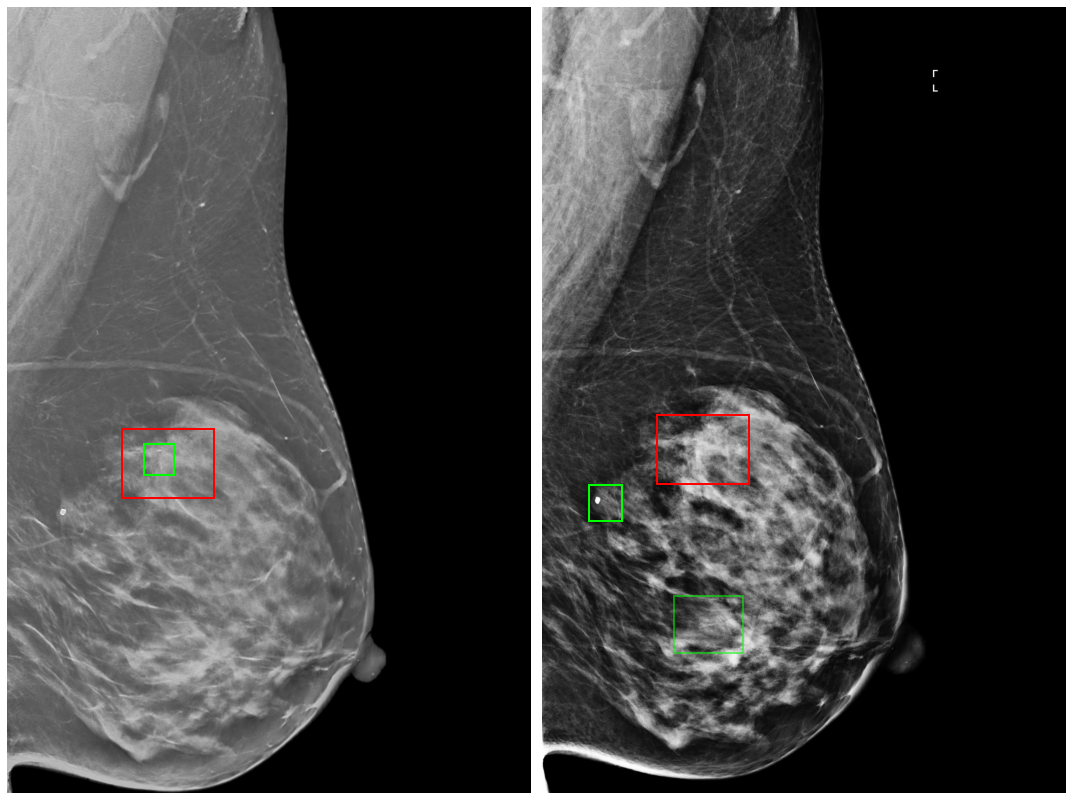} }}%
    \caption{False negative mis-classifications made by the FFDM model only. SM image is on the left of each set, FFDM to the right. }
    \label{fig-ffdm-fn}
  \end{minipage}
  \begin{minipage}[h]{\linewidth}
      {{\includegraphics[width=0.45\textwidth]{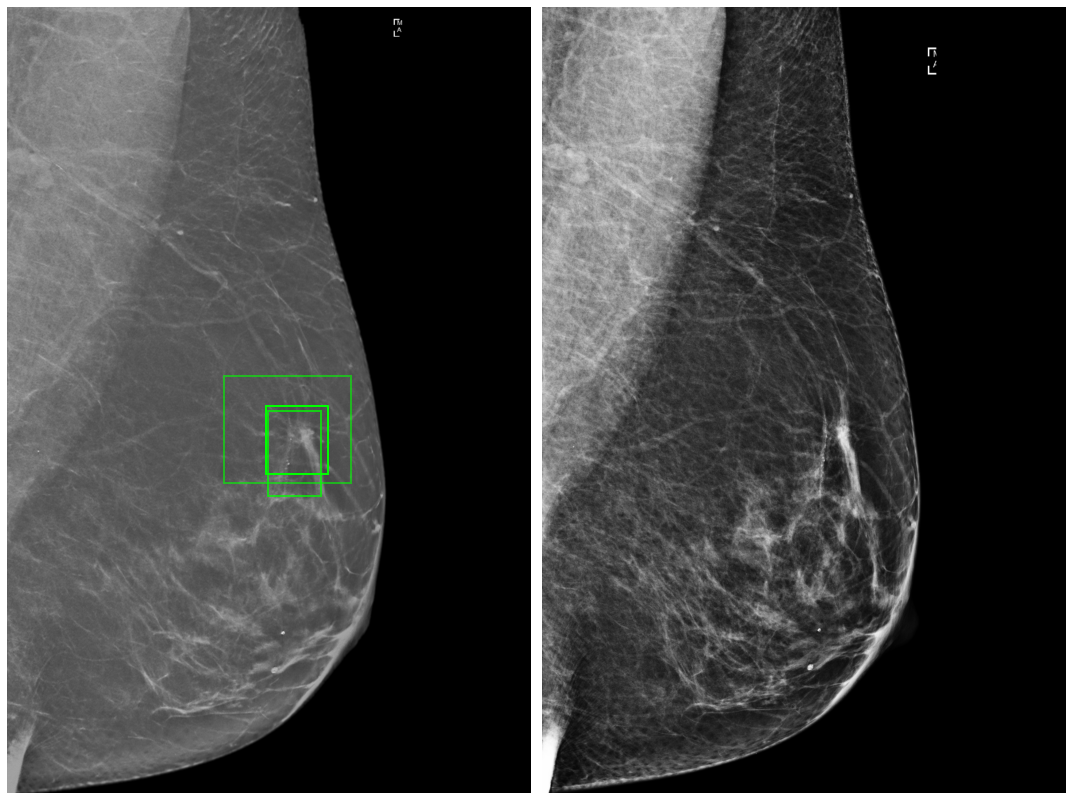} }}%
    \qquad
{{\includegraphics[width=0.45\textwidth]{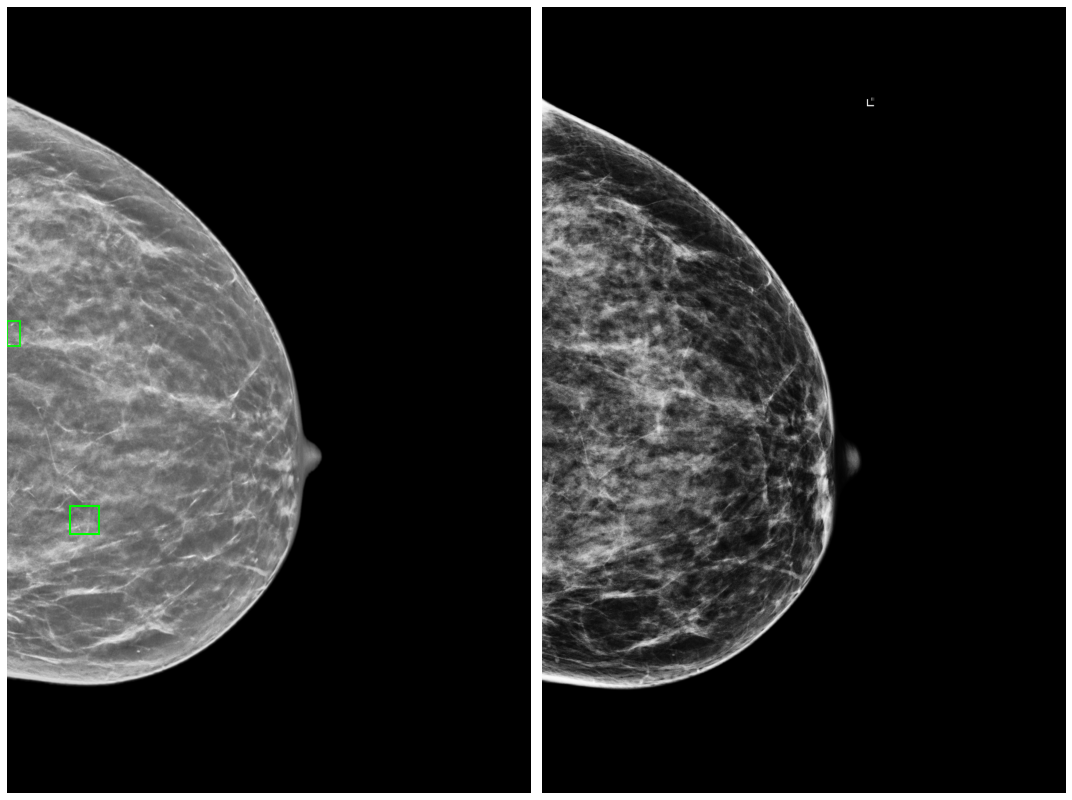} }}%
    \caption{False positive detections, only made by the model trained on SMs. SM image is on the left of each set, FFDM to the right.}
    \label{fig-cview-fp}
  \end{minipage}
  \begin{minipage}[h]{\linewidth}
      {{\includegraphics[width=0.45\textwidth]{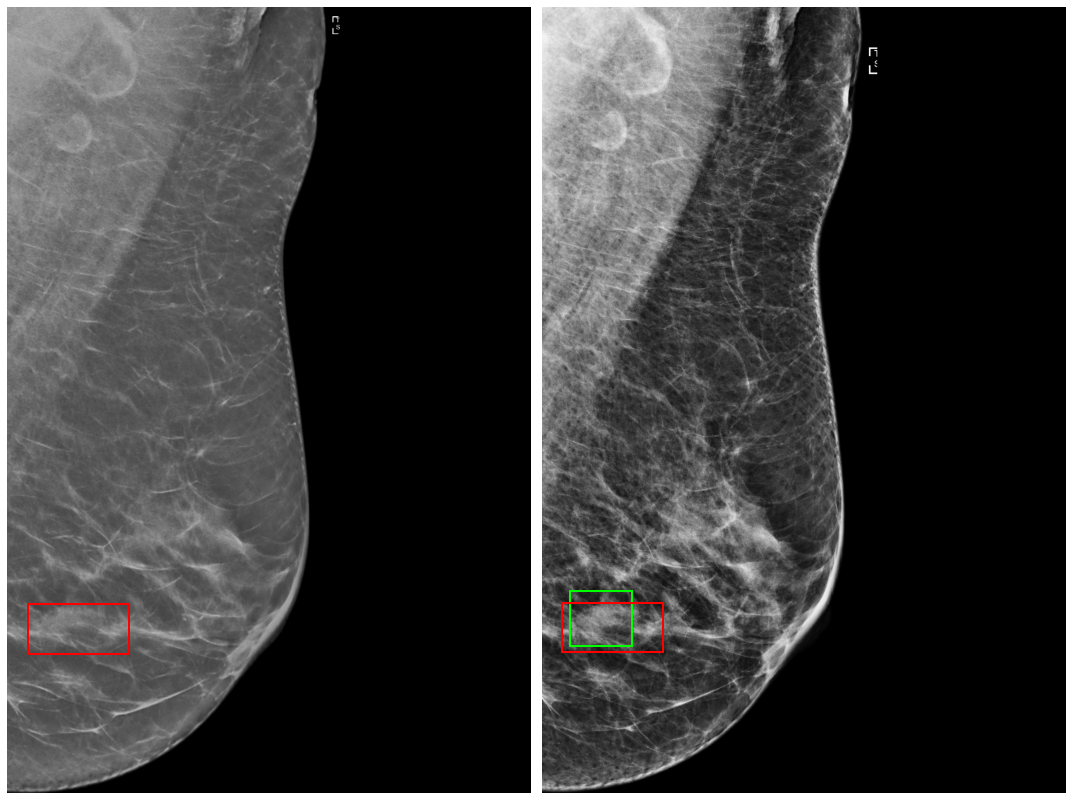} }}%
        \qquad
     {{\includegraphics[width=0.45\textwidth]{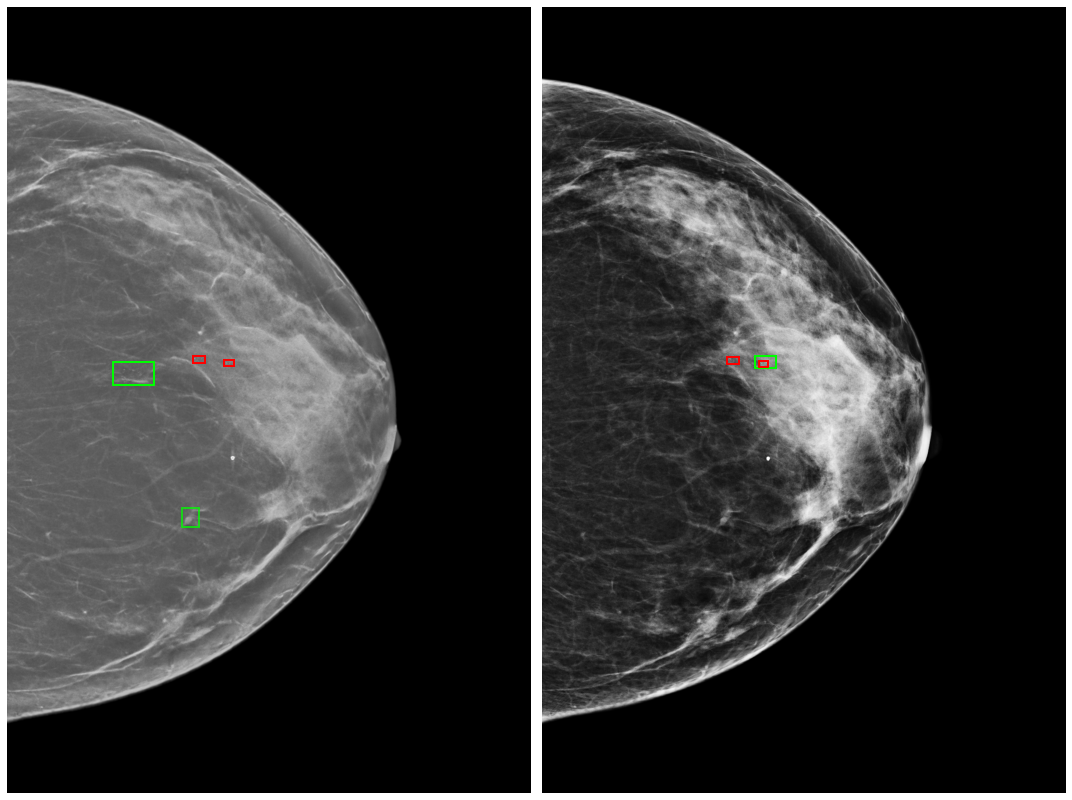} }}%
        \caption{Lesion is missed by the model trained on SMs but captured by the model trained on FFDMs. SM image is on the left of each set, FFDM to the right.}
        \label{fig-cview-fn}
    \end{minipage}
\end{figure}

%% file: miccai_submission.bbl
\begin{thebibliography}{10}
\providecommand{\url}[1]{\texttt{#1}}
\providecommand{\urlprefix}{URL }
\providecommand{\doi}[1]{https://doi.org/#1}

\bibitem{nyudbt}
Anonymous.: Anonymous dataset. Technical Report  (****)

\bibitem{assran2023selfsupervised}
Assran, M., Duval, Q., Misra, I., Bojanowski, P., Vincent, P., Rabbat, M., LeCun, Y., Ballas, N.: Self-supervised learning from images with a joint-embedding predictive architecture (2023)

\bibitem{Bandos2009}
Bandos, et~al.: Area under the free-response roc curve (froc) and a related summary index. Biometrics  \textbf{65}(1),  247--256 (Mar 2009)

\bibitem{cancers15102704}
Bobowicz, M., Rygusik, M., Buler, J., Buler, R., Ferlin, M., Kwasigroch, A., Szurowska, E., Grochowski, M.: Attention-based deep learning system for classification of breast lesions \& multimodal, weakly supervised approach. Cancers  \textbf{15}(10) (2023)

\bibitem{ciatto}
Ciatto, S., Houssami, N., Bernardi, D., Caumo, F., Pellegrini, M., Brunelli, S., Tuttobene, P., Bricolo, P., Fantò, C., Valentini, M.and~Montemezzi, S., Macaskill, P.: Integration of 3d digital mammography with tomosynthesis for population breast-cancer screening (storm): a prospective comparison study.  (2013)

\bibitem{geras_moy}
Geras, K.J., Mann, R.M., Moy, L.: Artificial intelligence for mammography and digital breast tomosynthesis: Current concepts and future perspectives. Radiology  \textbf{293}(2),  246--259 (2019), pMID: 31549948

\bibitem{dbt_review}
Gilbert, F.J., Tucker, L.~andYoung, K.C.: Digital breast tomosynthesis (dbt): a review of the evidence for use as a screening tool. Clinical radiology  \textbf{71},  141–50 (2016)

\bibitem{fastrcnn}
Girshick, R.B.: Fast {R-CNN}. CoRR  \textbf{abs/1504.08083} (2015)

\bibitem{GUR2012166}
Gur, D., Zuley, M.L., Anello, M.I., Rathfon, G.Y., Chough, D.M., Ganott, M.A., Hakim, C.M., Wallace, L., Lu, A., Bandos, A.I.: Dose reduction in digital breast tomosynthesis (dbt) screening using synthetically reconstructed projection images: An observer performance study. Academic Radiology  \textbf{19}(2),  166--171 (2012)

\bibitem{haas2013}
Haas, B.M., Kalra, V., Geisel, J., Raghu, M., Durand, M., Philpotts, L.E.: Comparison of tomosynthesis plus digital mammography and digital mammography alone for breast cancer screening. Radiology  \textbf{269}(3),  694--700 (2013), pMID: 23901124

\bibitem{maskrcnn}
He, K., Gkioxari, G., Doll{\'{a}}r, P., Girshick, R.B.: Mask {R-CNN}. CoRR  \textbf{abs/1703.06870} (2017)

\bibitem{resnet}
He, K., Zhang, X., Ren, S., Sun, J.: Deep residual learning for image recognition. CoRR  \textbf{abs/1512.03385} (2015)

\bibitem{hinton2015distilling}
Hinton, G., et~al.: Distilling the knowledge in a neural network (2015)

\bibitem{horvat2019}
Horvat, J.V., Keating, D.M., Rodrigues-Duarte, H., Morris, E.A., Mango, V.L.: Calcifications at digital breast tomosynthesis: Imaging features and biopsy techniques. RadioGraphics  \textbf{39}(2),  307--318 (2019), pMID: 30681901

\bibitem{dbtex_comp}
Konz, N., others.: {A Competition, Benchmark, Code, and Data for Using Artificial Intelligence to Detect Lesions in Digital Breast Tomosynthesis}. JAMA Network Open  \textbf{6}(2),  e230524--e230524 (02 2023)

\bibitem{shen2019}
L, S., LR, M., JH, R., E, F., R, M., W, S.: Deep learning to improve breast cancer detection on screening mammography. Sci Rep.  (2019), pMID: 31467326; PMCID: PMC6715802

\bibitem{li2021}
Li, H., Ye, J., et~al.: Application of deep learning in the detection of breast lesions with four different breast densities. Cancer medicine  \textbf{10}(14),  4994–5000 (2021)

\bibitem{coco}
Lin, T., Maire, M., Belongie, S.J., Bourdev, L.D., Girshick, R.B., Hays, J., Perona, P., Ramanan, D., Doll{\'{a}}r, P., Zitnick, C.L.: Microsoft {COCO:} common objects in context. CoRR  \textbf{abs/1405.0312} (2014)

\bibitem{focal_loss}
Lin, T.Y., et~al.: Focal loss for dense object detection (2017)

\bibitem{ssd}
Liu, W., Anguelov, D., Erhan, D., Szegedy, C., Reed, S.E., Fu, C., Berg, A.C.: {SSD:} single shot multibox detector. CoRR  \textbf{abs/1512.02325} (2015)

\bibitem{lu2019}
Lu, W., et~al.: Breast cancer detection based on merging four modes mri using convolutional neural networks. In: ICASSP 2019 - 2019 IEEE International Conference on Acoustics, Speech and Signal Processing (ICASSP). pp. 1035--1039 (2019)

\bibitem{proconsms2}
Melissa, D.: Synthesized mammography: Clinical evidence, appearance, and implementation  (2018)

\bibitem{cad_comp}
Niemeijer, M., Loog, M., Abràmoff, M.D., Viergever, M.A., Prokop, M., van Ginneken, B.: On combining computer-aided detection systems. IEEE Transactions on Medical Imaging  \textbf{30}(2),  215--223 (2011)

\bibitem{dbtex_comp_2}
Park, J., Shoshan, Y., Martí, R., others.: Lessons from the first dbtex challenge. Nature Machine Intelligence p. 735–736 (2021)

\bibitem{proconsms}
Ratanaprasatporn, et~al.: Strengths and weaknesses of synthetic mammography in screening. RadioGraphics  \textbf{37}(7),  1913--1927 (2017), pMID: 29131762

\bibitem{yolo}
Redmon, J., Divvala, S.K., Girshick, R.B., Farhadi, A.: You only look once: Unified, real-time object detection. CoRR  \textbf{abs/1506.02640} (2015)

\bibitem{yolo9000}
Redmon, J., Farhadi, A.: {YOLO9000:} better, faster, stronger. CoRR  \textbf{abs/1612.08242} (2016)

\bibitem{yolov3}
Redmon, J., Farhadi, A.: Yolov3: An incremental improvement. CoRR  \textbf{abs/1804.02767} (2018)

\bibitem{fasterrcnn}
Ren, S., He, K., Girshick, R.B., Sun, J.: Faster {R-CNN:} towards real-time object detection with region proposal networks. CoRR  \textbf{abs/1506.01497} (2015)

\bibitem{ribli}
Ribli, D., Horváth, A., Unger, Z., Pollner, P., Csabai, I.: Detecting and classifying lesions in mammograms with deep learning. Scientific reports  (2018)

\bibitem{overfeat}
Sermanet, P., Eigen, D., Zhang, X., Mathieu, M., Fergus, R., LeCun, Y.: Overfeat: Integrated recognition, localization and detection using convolutional networks (2013). \doi{10.48550/ARXIV.1312.6229}

\bibitem{skaane2019}
Skaane, P., Bandos, A.I., Niklason, L.T., Sebu\o{}deg\r{a}rd, S., \O{}ster\r{a}s, B.H., Gullien, R., Gur, D., Hofvind, S.: Digital mammography versus digital mammography plus tomosynthesis in breast cancer screening: The oslo tomosynthesis screening trial. Radiology  \textbf{291}(1),  23--30 (2019), pMID: 30777808

\bibitem{sobal2022joint}
Sobal, V., au2, J.S.V., Jalagam, S., Carion, N., Cho, K., LeCun, Y.: Joint embedding predictive architectures focus on slow features (2022)

\bibitem{spangler2011}
Spangler, M.L., Zuley, M.L., Sumkin, J.H., Abrams, G., Ganott, M.A., Hakim, C., Perrin, R., Chough, D.M., Shah, R., Gur, D.: Detection and classification of calcifications on digital breast tomosynthesis and 2d digital mammography: A comparison. American Journal of Roentgenology  \textbf{196}(2),  320--324 (2011), pMID: 21257882

\bibitem{efficientdet}
Tan, M., et~al.: Efficientdet: Scalable and efficient object detection  (2019)

\bibitem{wu2019}
Wu, M.C., et~al.: Multi-teacher knowledge distillation for compressed video action recognition on deep neural networks. In: ICASSP 2019 - 2019 IEEE International Conference on Acoustics, Speech and Signal Processing (ICASSP). pp. 2202--2206 (2019)

\bibitem{wu2020}
Wu, N., Phang, J., Park, J., et~al.: Deep neural networks improve radiologists' performance in breast cancer screening. IEEE transactions on medical imaging  \textbf{39}(4),  1184–1194 (2020)

\bibitem{yala2019}
Yala, A., Lehman, C., Schuster, T., Portnoi, T., Barzilay, R.: A deep learning mammography-based model for improved breast cancer risk prediction. Radiology  \textbf{292}(1),  60--66 (2019), pMID: 31063083

\bibitem{zhu2019}
Zhu, M., et~al.: Low-resolution visual recognition via deep feature distillation. In: ICASSP 2019 - 2019 IEEE International Conference on Acoustics, Speech and Signal Processing (ICASSP). pp. 3762--3766 (2019)

\end{thebibliography}
